%% file: bare_conf.tex
\documentclass[a4paper,fleqn]{cas-dc}
\usepackage{cite}

\usepackage{graphicx}
\usepackage{booktabs}
\usepackage[linesnumbered,ruled,vlined]{algorithm2e}
\usepackage{xcolor}
\usepackage{caption}
\usepackage{graphicx}
\usepackage{subfigure}
\usepackage{float} 
\usepackage[super]{nth}
\usepackage{multirow}
\usepackage{xspace} 
\usepackage{soul, color, xcolor}
\usepackage{bbding}
\usepackage{gensymb} 
\usepackage[super]{nth} 

\soulregister{\cite}7 
\soulregister{\citep}7 
\soulregister{\citet}7 
\soulregister{\ref}7 
\soulregister{\pageref}7 

\newcommand{\eg}{\textit{e}.\textit{g}.\xspace}

\newcommand{\od}{\textsf{LODAP}\xspace}
\newcommand{\eim}{\textsf{EIM}\xspace}

\newcommand{\incf}{\textsf{InCF}\xspace}

\begin{document}
\shorttitle{LODAP: On-Device Incremental Learning}
\shortauthors{Biqing Duan et~al.}

\title [mode=title]{LODAP: On-Device Incremental Learning Via Lightweight Operations and Data Pruning}



\author[1]{Biqing Duan}[orcid=0000-0001-0000-0000]
\cormark[1]
\ead{duanbiqing@stu.ynu.edu.cn}

\author[1]{Qing Wang}[orcid=0000-0001-0000-0000]
\cormark[1]
\ead{1746098836@qq.com}

\author[2]{Di Liu}[orcid=0000-0002-4365-2768]
\ead{di.liu@ntnu.no}

\author[1]{Wei Zhou}[orcid=0000-0002-5881-9436]
\ead{zwei@ynu.edu.cn}

\author[1]{Zhenli He}[orcid=0000-0002-7986-2222]
\ead{hezl@ynu.edu.cn}

\author[1]{Shengfa Miao}[orcid=0000-0003-1210-1135]
\ead{miaosf@ynu.edu.cn}

\affiliation[1]{organization={School of Software, Yunnan University},
                city={Kunming},
                country={China}}

\affiliation[2]{organization={Department of Computer Science, Norwegian University of Science and Technology},
                city={Trondheim},
                country={Norway}}

\cortext[cor1]{Equal contributors}

\nonumnote{This work is supported in part by NSFC under Grant 62362068, the National Key Research Development Program of China under Grant No.2024YFC3014300, and the Yunnan Province Major Science and Technology Project under Grant No.202302AD080006.}

\begin{abstract}
Incremental learning that learns new classes over time after the model's deployment is becoming increasingly crucial, particularly for industrial edge systems, where it is difficult to communicate with a remote server to conduct computation-intensive learning. 
As more classes are expected to learn after their execution for edge devices. In this paper, we propose \od, a new on-device incremental learning framework for edge systems.  
The key part of \od is a new module, namely Efficient Incremental Module (EIM).
\eim is composed of normal convolutions and lightweight operations. 
During incremental learning, \eim exploits some lightweight operations, called \textit{adapters}, to effectively and efficiently learn features for new classes so that it can improve the accuracy of incremental learning while reducing model complexity as well as training overhead.
The efficiency of \od is further enhanced by a data pruning strategy that significantly reduces the training data, thereby lowering the training overhead.
We conducted extensive experiments on the CIFAR-100 and Tiny-ImageNet datasets. Experimental results show that \od improves the accuracy by up to 4.32\% over existing methods while reducing around 50\% of model complexity. In addition, evaluations on real edge systems demonstrate its applicability for on-device machine learning. The code is available at https://github.com/duanbiqing/LODAP.
\end{abstract}


\maketitle

\input{introduction}
\input{related_work}
\input{method}
\input{experimental_results}

\input{conclusion}





%

\bibliographystyle{IEEEtran}
\bibliography{reference}

\end{document}

%% file: introduction.tex
\section{Introduction}
\label{sec:intro}
Deep neural networks (DNNs), particularly convolution neural networks (CNNs), have demonstrated remarkable performance across various fields, \eg, image classification and object detection \cite{bengio2021deep}. 
Since the training of CNNs is computation-intensive and CNNs themselves have been becoming increasingly complex to pursue higher accuracy, CNN models are typically trained on high-performance servers. 
Then, the well-trained model is deployed on a target platform, another server or an edge platform.
Such conventional \textit{training-deployment} paradigm may suffer from the inferior prediction accuracy when applying the well-trained model to a new context, in which some new data or categories are introduced after deployment. 
Thus, methods are proposed to address this issue, such as transfer learning or model adaption. 
In some cases, new training data becomes available or is introduced to the model \textit{incrementally}, necessitating \textit{incremental learning} (IL) capability \cite{he2011incremental}. 

An increasing number of CNNs have been gradually implemented on edge systems to enable local processing of sensitive and private data and ensure on-time inference, even with unreliable networks \cite{zhou2019edge}. 
The intersection of edge computing and artificial intelligence is commonly referred to as \textit{EdgeAI} or \textit{Edge Intelligence}. 
EdgeAI is deemed as the key element of ubiquitous AI, efficiently and effectively processing local data without expensive data communication. Some EdgeAI systems are required to learn new classes or tasks after their model deployment. 
However, data collected from edge systems are confidential or sensitive, and users are reluctant or forbidden to share these data with the model provider. 
As a result, on-device machine learning is emerging  as a promising solution \cite{dhar2021survey} and has been deployed on mobile devices\footnote{https://machinelearning.apple.com/research/recognizing-people-photos}. 
\textit{On-device incremental learning} is one of on-device machine learning methods which learns new classes on top of the deployed model \cite{wang2022efficient}\cite{de2021continual}.

On-device incremental learning (ODIL) poses challenges from both the perspectives of IL and edge systems. 
IL typically suffers from \textit{catastrophic forgetting} \cite{de2021continual} --
\textit{as the model learns new tasks, its performance on previous tasks rapidly deteriorates. }
The catastrophic forgetting is due to that the features learned for new classes overwrite the features of old classes \cite{de2021continual}.  
Hence, increasing the model capacity to learn new classes becomes a natural solution, but results in a more complex model \cite{yan2021dynamically}, which impedes its applicability on resource-constrained edge systems. 
Knowledge distillation is a widely used technique to mitigate catastrophic forgetting in IL \cite{li2017learning, castro2018end}. 
However, their performance is inferior in terms of achievable accuracy. 
Parameter freezing is proposed in \cite{wang2022efficient} to conduct ODIL. 
Parameter freezing utilizes redundant features/weights in the old model to learn new features. It can reduce the training overhead and mitigate catastrophic forgetting. 
However, as the number of incremental classes increases, its performance degrades significantly.


From the perspective of on-device ML, two factors contribute to the challenges of implementing ODIL on resource-constrained edge platforms: the\textit{ high model complexity} and the \textit{high training overhead}. 
\textbf{Model complexity}: Although new and effective model architectures have been proposed, CNN model design still adheres to the scaling law, aiming to increase parameters and capacities for higher accuracy. 
\textbf{Model training}: Model training is computationally intensive and memory-demanding, thus requiring substantial resources. 
Although IL only learns a few classes once, it still requires a large amount of diverse and high-quality data to train the model, which leads to high training overhead.
However, many edge platforms, such as those used in mobile edge computing, are subject to limited resources, memory, and battery power. 
The substantial training overhead can rapidly deplete an edge system's battery. 

In this paper, we propose a new ODIL framework for edge systems, dubbed \od, which deploys a lightweight module to address model complexity and data pruning to reduce training overhead.  
In \od, we propose a new lightweight operation to increase the capacity of the model which is used to learn features of the new classes without affecting the old classes. By doing so, we preserve the features of old classes while reducing both model size and training overhead. 
Additionally, we employ a fusion method after IL to integrate the newly learned features with the old ones, ensuring that the model size remains constant throughout the IL process. 
Then, 
inspired by the data pruning method \cite{paul2021deep}, we propose a simple yet effective data pruning method to prune the data during the IL phases. 
In particular, \od makes the following novel contributions: 

\begin{enumerate}
    \item We propose a new module for efficient IL, namely \textit{Efficient Incremental Module} (\eim), which consists of normal convolutional kernels and lightweight operations. 
 EIM only uses some lightweight operations, called \textit{adapters}, to learn new features for new tasks. 
 By means of \eim, we can reduce the model complexity and improve the effectiveness of IL. 
 To further reduce the increased complexity over the course of IL, we exploit \textit{structural reparameterization} to fuse the newly learned adapters into the main structure. 
        \item To further reduce the training cost, we propose a progressive data pruning method that is based on calculating the Error L2-Norm(EL2N) scores. 
        Our data pruning method prunes data with low EL2N scores. This method not only significantly reduces the training overhead, but also slightly improves the overall accuracy. This may provide a new lens for IL. 
    
    \item 
    Experimental results on CIFAR-100 and Tiny-ImageNet datasets show that \od can improve the accuracy by up to 4.32\% over the state-of-the-art methods while requiring only 50\% of model parameters. In addition, experiments on real edge platforms show \od can reduce the training cost on resource-constrained systems.
\end{enumerate}

The remainder of this paper is organized as follows: 
Section \ref{sec:related} discusses the related work. 
Section \ref{sec:prob} formulate the problem.
Section \ref{sec:prop} presents the details of \od. 
Section \ref{sec:exper} extensively evaluates \od in terms of accuracy and performance. 
Section \ref{sec:conc} concludes this paper. 



%% file: related_work.tex
\section{Related Work}
\label{sec:related}

\subsection{Incremental Learning}
The existing IL methods fall mainly into the following categories: 

\textbf{\textit{Replay-based}} methods, like \cite{rebuffi2017icarl, bang2021rainbow}, learn new classes with some exemplars of old classes. 
The key part in replay-based methods is to select some good examples of old classes.
\textsf{iCaRL} \cite{rebuffi2017icarl} selects some representative exemplars of a class using a strategy of the closest mean of exemplars. 
In contrast, \textsf{RM} \cite{bang2021rainbow} proposes a diversity-aware sampling method to select diverse exemplars as representative exemplars of old classes. 
\par

\textbf{\textit{Regularization-based}} methods usually deploy knowledge distillation \cite{li2017learning,douillard2020podnet,hou2019learning} and parameter isolation/reuse \cite{wang2022efficient,fernando2017pathnet,kirkpatrick2017overcoming}. 
Methods based on knowledge distillation train the new classes by learning knowledge from old models.
Parameter isolation approaches generally learn new tasks by only using part of the model so that the important features of old classes are not overwritten. 
For example, \textsf{PathNet} \cite{fernando2017pathnet}, \textsf{EWC} \cite{kirkpatrick2017overcoming}, and \textsf{EODI} \cite{wang2022efficient} propose different ways to freeze important weights for old classes and learn features for new classes with only those unimportant weights. 
Among them, \textsf{EODI} is the only one targeting on-device learning.

\textbf{\textit{Structure-based}} methods typically address catastrophic forgetting by increasing the capacity of the model. \textsf{DER} \cite{yan2021dynamically} learns new knowledge by keeping the learned parameters fixed and then expanding the model. However, the main problem is that the model keeps growing as the number of categories increases. \textsf{SSRE} \cite{zhu2022self} introduces a dynamic incremental module to increase a model's capacity.

\textbf{\textit{Discussion}}: In the context of on-device learning, the aforementioned methods pose different challenges. Replay-based methods typically require additional storage space for old class images, which can be unfriendly to resource-constrained devices. Recently, some prototype-based methods have been proposed to address this issue, such as \cite{zhu2021prototype}. 
This is because the model remains the same, fewer irrelevant weights can be exploited as more classes are added.  
Structural-based methods often exhibit superior accuracy, but result in more complex models and higher training overhead, making them unsuitable for on-device learning on resource-constrained devices. 

\subsection{Data Pruning}
Data pruning, by removing redundant or low-value samples from the training set, has become one of the key techniques for improving deep learning efficiency while maintaining model performance and reducing computational costs. 
Toneva et al. \cite{toneva2018empirical} pioneered the concept of forgetting events and the forgetting count metric, quantifying the importance of data by counting the frequency of misclassified samples during training.
SVP \cite{coleman2019selection} proposed using a lightweight proxy model instead of a complete model for data selection and designed a greedy algorithm based on the proxy model to quickly filter the most valuable subset of samples for the target task, thus achieving data pruning. 
CRAIG \cite{mirzasoleiman2020coresets} implemented a greedy algorithm based on submodel optimization, maximizing the feature coverage of the coreset from a geometric perspective.
Paul et al. \cite{paul2021deep} proposed the GraNd score metric and further introduced the norm of the error vector, which dynamically filters samples using the gradient amplitude at the beginning of training and then restarts training using the pruned data.YOCO \cite{he2024you} proposed two principled pruning rules, combining sample influence and diversity to design an efficient and scalable pruning method.

While current data pruning methods have shown potential in different scenarios, they still exhibit certain limitations. For example, some methods have high computational complexity during data pruning, and the computational costs may even exceed the savings achieved by training on the pruned dataset. This contradicts the original intention of data pruning. Future research should focus on reducing computational costs, improving the generality and robustness of methods, and exploring more efficient techniques for data pruning.


%% file: method.tex
\section{\od Framework}
\label{sec:prop}
\begin{figure*}[ht]
    \centering
    \includegraphics[width=0.9\linewidth]{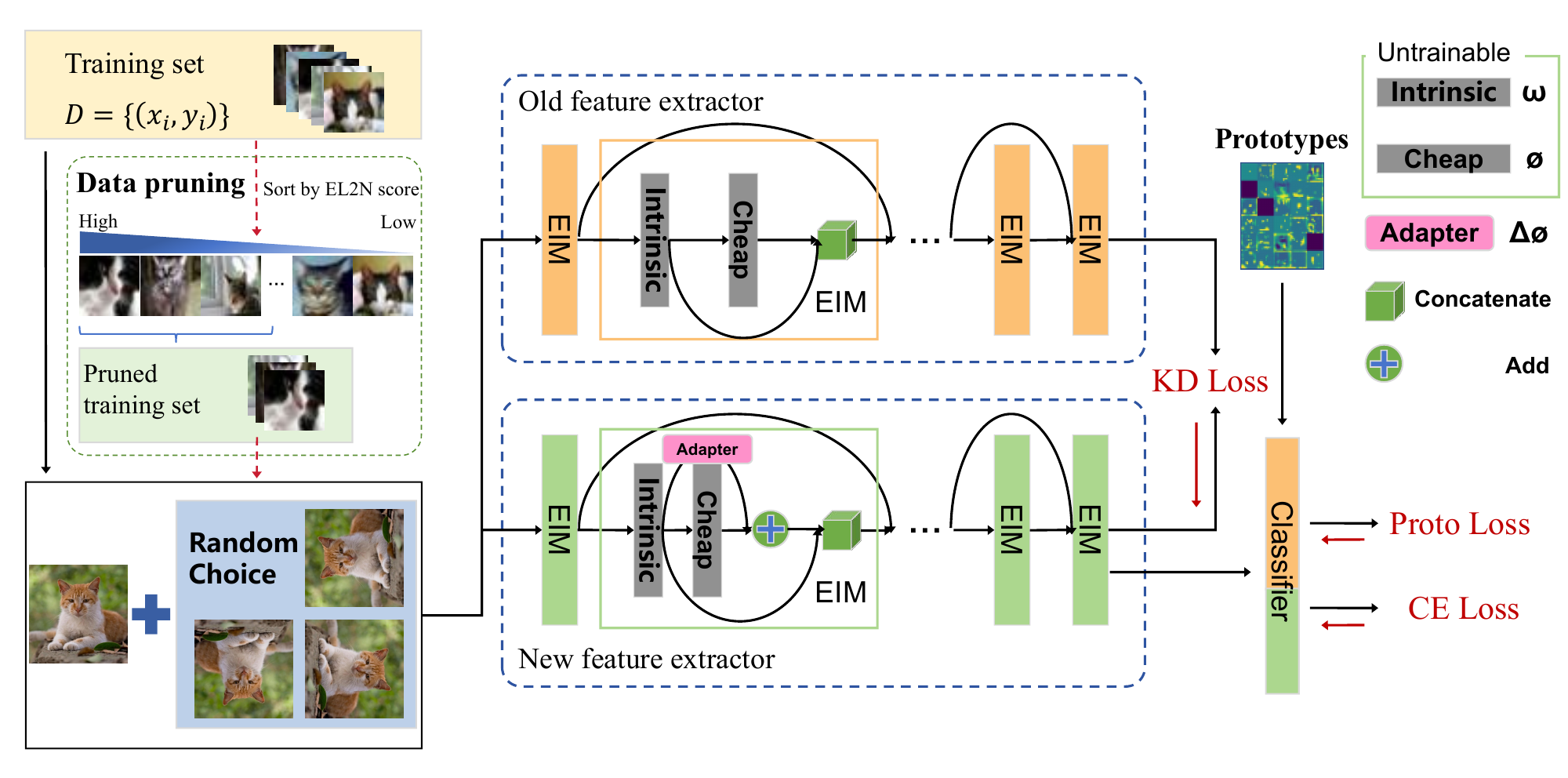}
    \caption{The overview of \od.}
    \label{framework}
\end{figure*}

This section first formulates our problem and presents the details of \od. 
\od targets edge systems which are usually subject to limited computing and memory resources. 
The overview of \od is in Fig. \ref{framework}, where \textit{adapters} in \eim are used to learn features of new tasks and features of old tasks are preserved using intrinsic features and some lightweight operations. 

\input{Problem}



\subsection{Efficient Incremental Module}
\label{propose:EIM}

To conduct an on-device learning for resource-constrained systems, the method should not increase the model's complexity while effectively and efficiently learning new features of new classes. 
We propose \eim to efficiently learn features of new classes. 
\eim is inspired by GhostNet \cite{han2020ghostnet}, 
where 
expensive convolutions are only used to derive some `\textit{intrinsic}' features and other similar features are computed using intrinsic features and \textit{cheap operations}.
By doing so, the model's complexity can be significantly reduced.

\eim in \od exploits some lightweight operations, called \textit{adapters}, to learn distinguished features for new classes. 
Then, the features of old classes can be preserved with the intrinsic features and some cheap features\footnote{To distinguish, when we refer to cheap operations/features afterwards, we mean these reserved for the old classes and are not trained during IL.} while features for new classes are learned in an efficient way. 
\eim has two types of operations which play different roles: 
\subsubsection*{Normal Convolutional Operations}
The normal convolutional operations generate the normal features within EIMs, also namely \textit{intrinsic features} (\textsf{InTFs}), which are obtained in prior to the phase of IL and not updated during IL. 
Formally, for a given input $l\in \mathbb{R}^{c \times h \times w}$, where $c$ is the number of channels and $h$ and $w$ are the height and width of the input, respectively. \eim generates the intrinsic features of old classes using Eq.~(\ref{eq:intrinisc}):
\begin{equation}
\label{eq:intrinisc}
f^{\prime}=Conv\left(l ; \omega\right)+\mathrm{b}
\end{equation}
where $Conv$ indicates a convolutional operation, $\mathrm{\omega} \in \mathbb{R}^{\mathrm{m} \times \mathrm{c} \times \mathrm{k} \times \mathrm{k}}$ is the convolution kernel, 
where $\mathrm{m}$ and $\mathrm{c}$ are the number of output and input channels, respectively, and $\mathrm{k} \times \mathrm{k}$ is the size of the convolution kernel.
$f^{\prime} \in \mathbb{R}^{\mathrm{m} \times \mathrm{h}^{\prime} \times \mathrm{w}^{\prime}}$ is the output feature map of the convolutional layer, where $\mathrm{h}^{\prime}$ and $\mathrm{w}^{\prime}$ are the height and width of the output feature map, respectively, and $b$ is a bias. 
\textit{The main task of \textsf{InTFs} is to preserve important features for the model, and more features can be derived using \textsf{InTFs} later.}

\subsubsection*{Lightweight Operations}
Lightweight operations generate more features using \textsf{InTFs} with fewer computations. 
We divide the lightweight operations within each \eim into two parts: one part, namely \textit{Adapters}, used to learn features for new classes, and another part used for old classes to have more distinguished features. 
We call the features \textit{incremental features} (\incf{s}), if they are learned by \textit{adapters} for new classes. \incf{s} are computed as: 
\begin{equation}
f_{Inc}=Adapter\left(f^{\prime} ;\Delta\phi\right) + \mathrm{b^{\prime}} 
\label{fg}
\end{equation}
where $Adapter$ learns \incf{s} for new tasks. 
$f^{\prime} \in \mathbb{R}^{\mathrm{m} \times \mathrm{h}^{\prime} \times \mathrm{w}^{\prime}}$ is the intrinsic features, and $\Delta\phi \in \mathbb{R}^{m* (\mathrm{s-1}) \times k' \times k'}$ is the kernel weights of an \textit{adapter}. 
For each channel of the intrinsic features, the \textit{adapter} uses a $k'\times k'$ kernel to conduct convolution operation with \textbf{only one} of $m$ channels to generate $s-1$ new \incf{s}\footnote{The lightweight operation does not convolve along the depth and thus significantly reduces the computation.}, where $k'=1$. 
Thus, an \textit{adapter} generates $m*(s-1)$ \incf{s}.
$f_{Inc} \in \mathbb{R}^{\left(\mathrm{m*\left(\mathrm{s-1}\right)}\right) \times \mathrm{h}^{\prime} \times \mathrm{w}^{\prime}}$ represents the generated \incf feature maps, $s$ is the hyper-parameter with $s*m=n$, and $n$ is the number of the output feature maps from the \eim.
In \od, we set $s=2$, i.e., each channel in an intrinsic feature generates one \incf. By using such lightweight operations, the number of computations can be reduced by half seen in Sec. \ref{sec:exper}.  

Cheap features that are reserved for old classes and not updated during IL are also computed using Eq.~(\ref{fg}) but with a $3 \times 3$ kernel. We use $\phi$ to indicate the weights of a layer's lightweight operations.


\subsubsection*{Adapter Fusion}
Applying \eim to IL is similar to increasing model's capacity, because new adapters are added. Although adapters have fewer parameters and low training overhead, it may lead to to large memory footprint as the number of IL steps increases. 
This may unfortunately impede the implementation of \eim-based models on resource-constrained systems. 
Thus, in \od, we propose to fuse the newly \incf{s} into the existing cheap features using structural re-parameterization which was first introduced in RepVGG \cite{ding2021repvgg} to increase accuracy and boost inference and was later exploited in IL \cite{zhu2022self}.

Adapters are only different from the reserved cheap operations in terms of the size. Thus, we use padding to make adapters the same size as cheap features. Let $\Delta\phi$ be the weight of an adapter, and the process is as follows:
\begin{equation}
\phi_t = \text{zero-padding}\left(\Delta\phi\right) + \phi_{t-1}
\end{equation}
where $\phi_t \in \mathbb{R}^{\mathrm{m}*\left(\mathrm{s-1}\right) \times k' \times k'}$ indicates the fused weights of a cheap feature at the t\textsuperscript{th} incremental step, and $\phi_{t-1}$ indicates the parameter of the cheap feature before fusion \footnote{$\Delta\phi$ can be negative or positive, so fusing is realized through addition instead of average}. By means of adapter fusion, the complexity of the model will not be increased after an IL step, making it suitable for memory-limited devices.

\subsection{Prototypes For Old Classes}
\label{proposed:proto}
IL is usually required to reserve a collection of training data for old classes and then mix them with the training data of new classes to prevent the model from forgetting old classes. 
However, reserving images poses a memory challenge for resource-constrained systems. 
As the number of old classes is gradually increased over the course of IL, the reserved images (e.g., from 20 to hundreds of images per class \cite{zhu2021prototype}) will result in a large memory occupation which can affect the efficiency of many edge accelerators. 

In \od, instead of storing images for old classes, we deploy the \textit{prototype} paradigm as \cite{zhu2021prototype}, where each class in the model has a \textit{prototype} that is a feature map obtained from the model. 
The merits of using \textit{prototypes} are twofold in \od: (1) \textit{prototypes} are more memory efficient than storing images; (2) \textit{prototypes} provide better protection to data privacy. For instance, many models are provided by vendors who may use some private data to train the model, and these data cannot be provided to users. 
In this case, \textit{prototypes} allow to retain some knowledge for old classes without sharing data. 
Taking CIFAR-100 as an illustrative example, assume 20 images per class and \textsf{uInt8} format, the memory allocation for a single class amounts to $(20 \times 3 \times 32 \times 32)/ 1024 =61$ kB. In contrast, \od uses the input of the final classifier to generate a prototype, where all training inputs' feature maps before classifier are calculated on average to derive the prototype for one class. The size of the feature map is $1\times 512$ and the feature map is stored in \textsf{Float32}. Hence, it only needs $512 \times 4\mathrm{B} / 1024= 2$ kB, 30X smaller. 

\subsection{Data Pruning}
For resource-constrained edge system devices, reducing ODIL training overhead is critical. \eim achieves this by lowering model complexity. 
Another factor contributing to high training overhead is data volume. 
Typically, a model requires a large volume of diverse, high-quality data for training. 
Recent work reveals that not all data are equally important for training an effective model \cite{paul2021deep}. 
Training with difficult data can improve model accuracy and reduce training overhead. 
We incorporate data pruning into \od to further reduce training overhead.  
We propose a simple yet effective training strategy based on EL2N scores. This approach quantifies the importance of data in the early training stages and progressively prune the unimportant data through the IL procedure. 
Pruning irrelevant data reduces the amount of required training data, thus significantly lowering training overhead. 
Additionally, data pruning not only lowers training overhead for our \od, but also improves the model's accuracy (See results in Section \ref{sec:exper}).

Define the dataset $D=\{(x_{i},y_{i})\}_{i=1}^{N}$, where $x_{i}$ is the $i^\text{th}$ example, and $y_{i}$ is its corresponding true label. The EL2N score for each data example $(x_{i},y_{i})$ is defined as follows:
\begin{equation}
EL2N(x_i,y_i)=||p(w_t,x_i)-\hat{y}_i||_2
\end{equation}
where $p(w_t,x_i)$ is the predicted probability of each example at time $t$ under the weight $w$, and $\hat{y}_i$ represents the one-hot encoded vector of the true label $y_i$. The EL2N score is obtained by calculating the L2-Norm between the predicted probability and the true label of the training example. 
\textit{The higher the EL2N score of an input sample, the more difficult it is for the model to learn. 
Therefore, samples with higher EL2N scores should be prioritized for training. }

Based on the EL2N score, we propose a progressive data pruning method for reducing the training data during the incremental learning stage. 
We assume that sufficient training data have been collected for the incremental classes. Increment learning begins with a standard training phase, where all available data are used to train the model. 
This standard training spans the first $n$ epochs and calculates the gradients required for the EL2N scores.
After the first training phase, the EL2N score of each sample has been calculated, and the data pruning can be conducted from the $(n+1)^\text{th}$ epoch. 
Samples are sorted according to their EL2N scores within each class, and a certain percentage of samples with lower scores are pruned from each class, and the remaining examples are used for subsequent training epochs. 
Reducing the number of training samples can significantly decrease the training overhead. 
The training after data pruning, apart from the reduction in the number of data examples, all other parameters and settings remain the same as in the standard training. We conducted an ablation study in Section \ref{exp:ablation} to evaluate the effectiveness of this training method.

\od requires reserving a prototype for each class as described in Section \ref{proposed:proto}. At the end of an incremental training process, we use the pruned data to generate the required prototype for each newly learned class. 

\subsection{Training}
\label{proposed:traning}
This section explains how to train a model in \od. 
\textit{The objective of the training is to learn sufficient features for new classes while avoiding forgetting old classes.} 
Our training method is similar to those for prototype-based IL, like \cite{zhu2022self}\cite{zhu2021prototype}. 
The old model from the last step also joins the training procedure as shown in Fig. \ref{framework}.
The training consists of three parts, and each part has its own loss function, aggregating to the final training loss for \od: a new-class loss $L_{ce}$, a knowledge distillation (KD) loss $L_{kd}$, and a prototype loss $L_{proto}$. 


\subsubsection*{New-Class}
This part is just like a normal classification model training. 
Its main purpose is to train the adapter in each \eim module with the data of new classes such that the new features are learned to predict the new classes. 
New-Class updates both the backbone and classifier of the model, but it only updates InCFs. \od adopts the common cross-entropy loss for New-Class, and the loss function is as follows:
\begin{equation}
\small
L_{ce}=-\sum_{\left(x_{i}, y_{i}\right) \in \mathrm{X}_{\mathrm{t}}} y_{i} \log \left(q_{i}\right)
\vspace{-0.5em}
\end{equation}
where $x_i$ is the arbitrary input data of $X_t$, the data of the $\text{t}^\text{th}$ training stage, $y_i$ is its corresponding label, and $q_i$ is the softmax output probability of $x_i$ from the model's classifier. 

\subsubsection*{Knowledge Distillation}
We use the old model as the teacher model to distill knowledge to the new model. 
KD only applies to the backbone during training. It transfers the \textit{knowledge} of the model, especially the feature of old classes, to the new model so that it can mitigate catastrophic forgetting for the backbone. The loss function is as follows:
\begin{equation}
L_{kd}=\left\|\mathcal{F}\left(X_t ; W, \Phi_{t-1},\Delta\Phi\right)-\mathcal{F}\left(X_t ; W, \Phi_{t-1}\right)\right\|
\end{equation}
where $\mathrm{W}$ represents the weights of all intrinsic features and other layers, excluding the cheap operations and adapters.
$\Phi_{t-1}$ represents the weights of all cheap operations, while $\Delta\Phi$ contains the weights of all adapters. 
$\left\| \cdot \right\|$ is L2-Norm.


\subsubsection*{Prototype}
Prototypes are only applied to the classifier. 
The main objective of prototypes is to calibrate the classifier for the old classes with new features. 
We also use a cross-entropy loss for it. 
The loss function is as follows:
\begin{equation}
L_{proto}=\sum_{\left(p,y\right) \in \mathcal{P}} L_{ce}\left(G\left(p ; \Theta_{t}\right), y\right)
\end{equation} 
where $p$ and $y$ represent the class prototype and its corresponding class label, respectively, and $\Theta_t$ represents the parameter of classifier $G$. 



\subsubsection*{The Final Training Loss} By combining the above-mentioned loss functions, we obtain the final loss function of \od as follows:
\begin{equation}\label{eq:loss}
L=L_{ce}+\gamma * L_{kd}+\lambda * L_{proto} 
\end{equation} 
where $\gamma$ and $\lambda$ represent coefficients of distillation loss and prototype loss, respectively. An ablation study on the effectiveness of this loss function is provided in Section \ref{exp:ablation}.

\subsection{Data Augmentation Strategy}
\label{proposed:data}
Data augmentation has proven its effectiveness in IL by learning more discriminative features \cite{zhu2021prototype}, where input images for new classes are manipulated to generate more data for training. 
\textsf{PASS} in \cite{zhu2021prototype} generates three rotated images  (90\degree, 180\degree, and 270\degree) for each training input. However, this results in a significant increase in memory usage. 
\od proposes a simple but effective data augmentation technique. 
Instead of generating three rotated images, we randomly generate one rotated image. 
Then, our method can reduce the memory requirement as well as training overhead. 
To evaluate the effectiveness of this data augmentation approach, we conduct an ablation study in Section \ref{exp:ablation}. 



%% file: Problem.tex
\subsection{Problem Statement}
\label{sec:prob}
The task of class incremental learning is to learn a feature extractor ($\mathcal{F}$) and a unified classifier ($G$) as the number of classes gradually increases over time. 
The model should not only learn the new class knowledge, but also not forget the previously learned classes. 
Formally, let $t \in \{0, 1, ..., T\}$ denote an incremental learning stage, where 0 is the initial stage, and 1 to $T$ represent incremental stages.
$\mathcal{D} = \cup^T_{t=0}\{X_{t}, Y_{t}\}$ denotes all training data the model may learn, where $X_t$ is the t\textsuperscript{th} incremental input data and $Y_t$ is its corresponding class labels. At the t\textsuperscript{th} stage, the model training data is $\mathcal{D}_t$, which includes all data the model has received right before the IL starts. Our goal at each stage is to minimize the total training loss for all classes, new and old classes,
\begin{equation}\label{1}
\arg \min _{W_t, \theta_t} L\left(G\left(\mathcal{F}\left(X_t ; W_t\right) ; \Theta_t\right), Y_t\right)
\end{equation}
where $W_t$ and $\Theta_t$ represent the parameters of $\mathcal{F}$ and G at the t\textsuperscript{th} incremental phase, respectively. $L$ is a loss function defined in a specific task, e.g., cross-entropy loss for the classification task.

%% file: experimental_results.tex
\section{Experimental Results}
\label{sec:exper}

\begin{table*}[htbp]
    \centering
    \small
    \begin{tabular}{c|c|c|c|c|c|c|c}
    \hline
        \multirow{2}{*}{Method} &\multirow{2}{*}{\#Param(M)}&\multirow{2}{*}{\#FLOPs(G)} & \multicolumn{2}{c|}{CIFAR-100}&\multicolumn{3}{c}{Tiny-ImageNet}\\ \cline{4-8}
        ~ & ~ & ~ & P=5 & P=10 & P=5 & P=10 & P=20 \\ \hline
        LwF-MC\cite{li2017learning} & 11.2  & 0.56  & 45.93  & 27.43  & 29.12  & 23.10  & 17.43  \\ 
        iCaRL-CNN\cite{rebuffi2017icarl} & 11.2  & 0.56  & 51.07  & 48.66  & 34.64  & 31.15  & 27.90  \\ 
        iCaRL-NCM\cite{rebuffi2017icarl} & 11.2  & 0.56  & 58.56  & 54.19  & 45.86  & 43.29  & 38.04  \\ 
        EEIL\cite{castro2018end} & 11.2  & 0.56  & 60.78  & 56.05  & 47.12  & 45.01  & 40.50  \\ 
        EWC\cite{kirkpatrick2017overcoming} & 11.2  & 0.56  & 24.48  & 21.20  & 18.80  & 15.77  & 12.39  \\ 
        UCIR\cite{hou2019learning} & 11.2  & 0.56  & 63.78  & 62.39  & 49.15  & 48.52  & 42.83  \\ 
        EODI\cite{wang2022efficient} & 11.2  & 0.56  & 63.01 & 59.85  &  48.97 & 46.08  & 41.95  \\ 
        PASS\cite{zhu2021prototype} & 11.4  & 0.56  & 63.47  & 61.84  & 49.55  & 47.29  & 42.07  \\ 
        SSRE\cite{zhu2022self} & 12.4  & 0.62  & 65.88  & 65.04  & 50.39  & 48.93  & 48.17  \\ 
    \od w/o DP & \textbf{5.8{\color{red} \scriptsize{-5.4}}}  & \textbf{0.28{\color{red} \scriptsize{-0.28}}}  & \textbf{67.21{\color{red} \scriptsize{+1.33}}}  & \textbf{65.97{\color{red} \scriptsize{+0.93}}} & \textbf{53.54{\color{red} \scriptsize{+3.15}}}  & \textbf{53.01{\color{red} \scriptsize{+4.08}}}  & \textbf{52.38{\color{red} \scriptsize{+4.21}}}  \\ 
        \od w DP & \textbf{5.8{\color{red} \scriptsize{-5.4}}} & \textbf{0.28{\color{red} \scriptsize{-0.28}}} & \textbf{67.41{\color{red} \scriptsize{+1.53}}} & \textbf{66.33{\color{red} \scriptsize{+1.29}}} & \textbf{53.56{\color{red} \scriptsize{+3.17}}}  & \textbf{53.02{\color{red} \scriptsize{+4.09}}}  &  \textbf{52.49{\color{red} \scriptsize{+4.32}}} \\ \hline
    \end{tabular}
    \caption{The average incremental accuracy and model complexity of different methods. }
    \label{average accuracy}
\end{table*}

\subsection{Experimental Setting}
\label{exp_set}

\noindent\textbf{Datasets:} 
Similar to the previous methods, we use the CIFAR-100 and Tiny-ImageNet datasets.  CIFAR-100 contains 100 categories, each with 600 RGB images of $32\times32$, where each class has 500 images for training and 100 images for test. Tiny-ImageNet has 200 categories, where each class has 550 images of $64\times64$ for training and 50 images for test.

\noindent\textbf{Architecture:}
ResNet-18 is selected as the backbone network for our experiment because we target a resource-constrained platform. Moreover, other reference methods \cite{wang2022efficient, zhu2021prototype, zhu2022self} also use ResNet-18 as their experimental architecture. 
Our approach just replaces convolutional layers with the new \eim modules. 
\noindent\textbf{Training:} 
 During the initial training stage, we randomly select half of the classes from the dataset and train the model with the selected classes. 
 The initial learning rate is set to 0.001, and the learning rate is decreased by a factor of 0.1 every 45 epochs. 
 We train the initial model for 100 epochs. In the incremental learning stage, we use the same optimizer and learning rate update strategy. 
 The initial learning rate is set to 0.0002, and the training is performed for 60 epochs.
 We perform data pruning after 20 epochs of training in each incremental phase. During the subsequent 40 epochs, we retain \textbf{30\% }of the examples for the CIFAR-100 dataset and \textbf{50\%} of the examples for the Tiny-ImageNet dataset.
 $\lambda$ and $\gamma$ in Eq.~(\ref{eq:loss}) are both set to 10. 
 In the experiments, \textsf{P} indicates the model learns the rest of classes by P steps, e.g., 5 phases for CIFAR-100 means it takes 5 IL steps, and every step it learns 10 classes.\par

\begin{figure}[htbp]
    \centering
    \begin{subfigure}
        \centering
        \includegraphics[width=0.65\columnwidth]{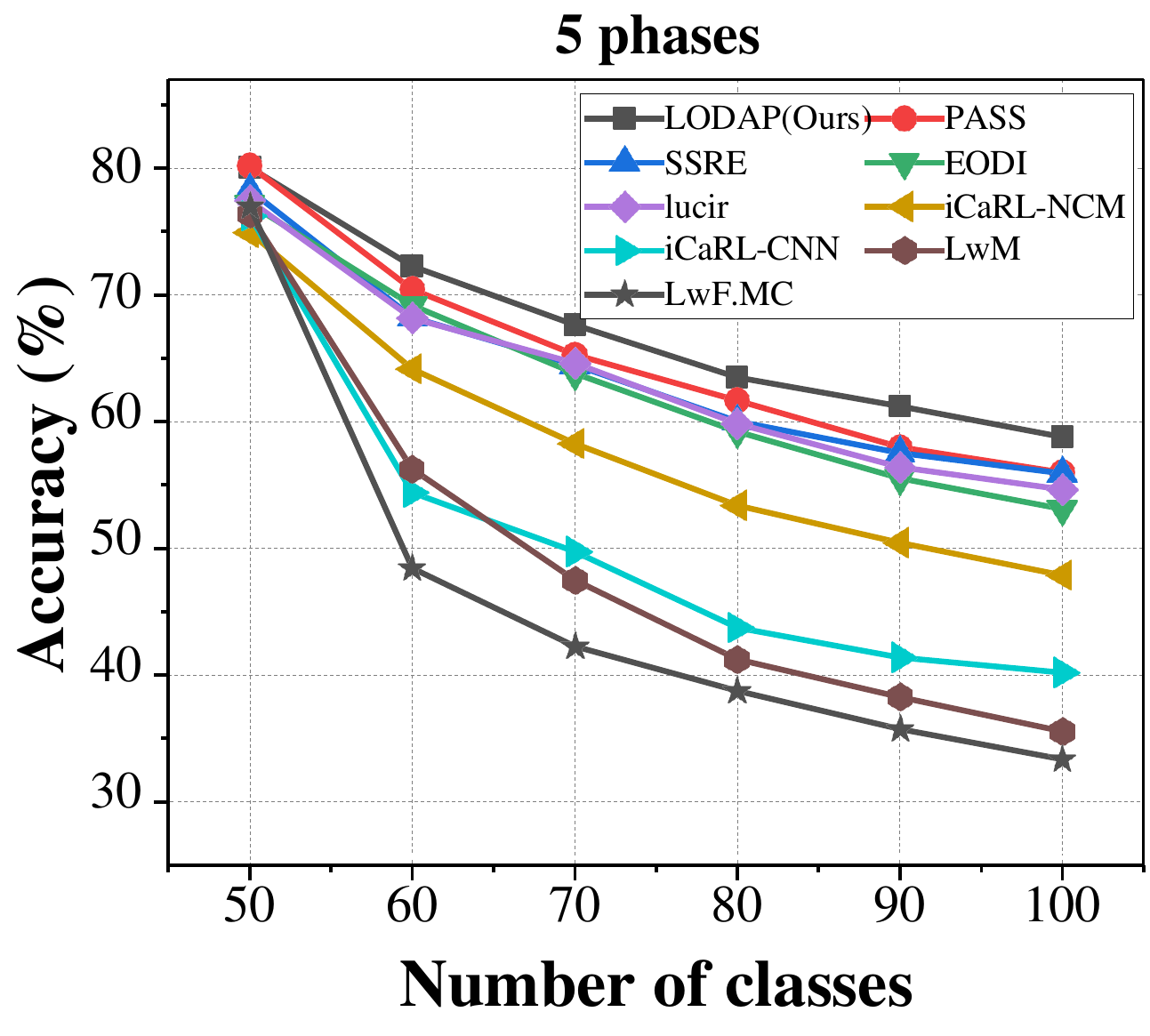}
        \label{5-cifar}
    \end{subfigure}
    \begin{subfigure}
        \centering
        \includegraphics[width=0.65\columnwidth]{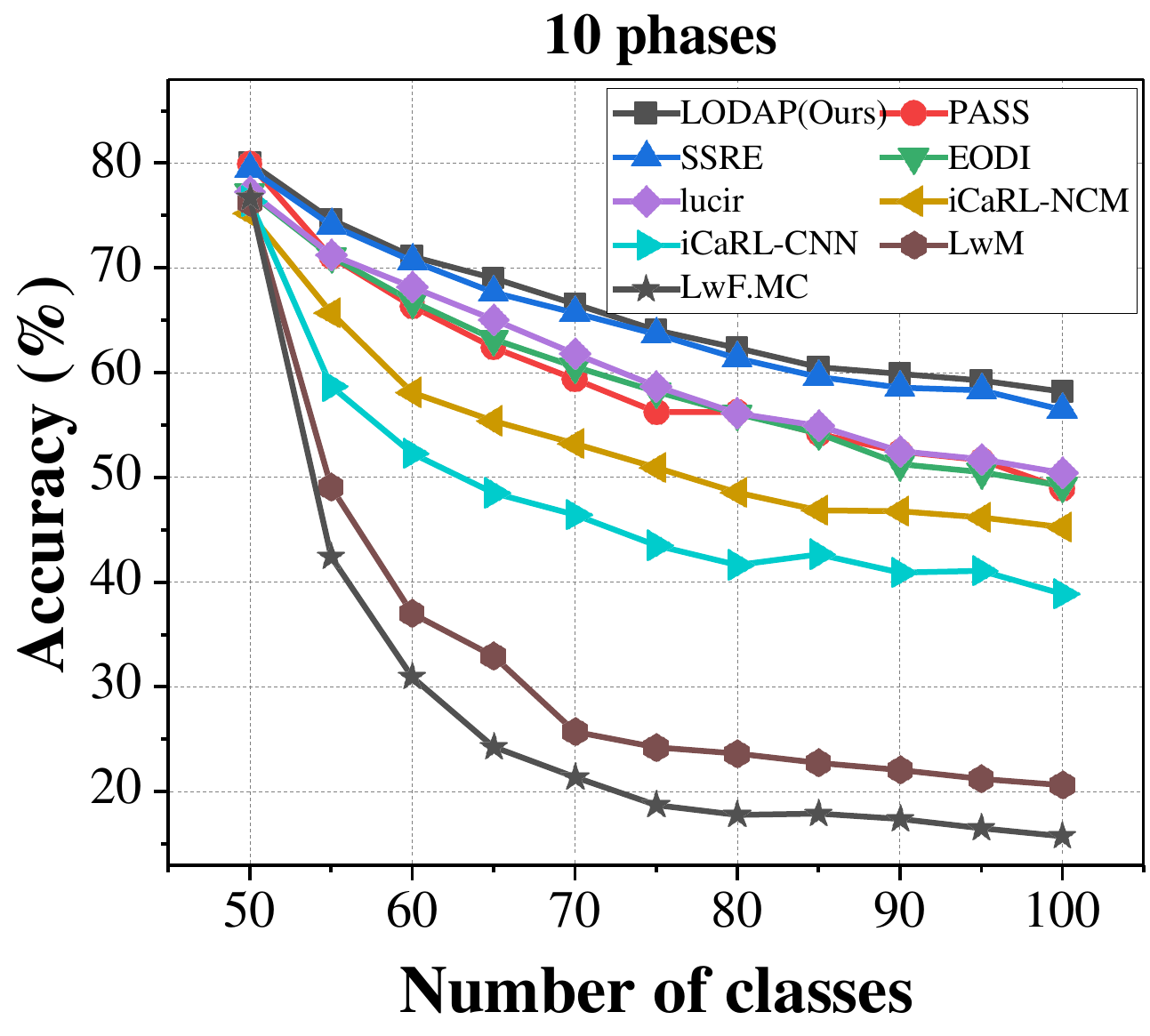}
        \label{10-cifar}
    \end{subfigure}
	\caption{The accuracy variations on CIFAR-100.}
        \label{CIFAR100}
\end{figure}

\begin{figure*}[htbp]
    \centering
    \begin{minipage}[t]{0.3\linewidth}
        \centering
        \includegraphics[width=\textwidth]{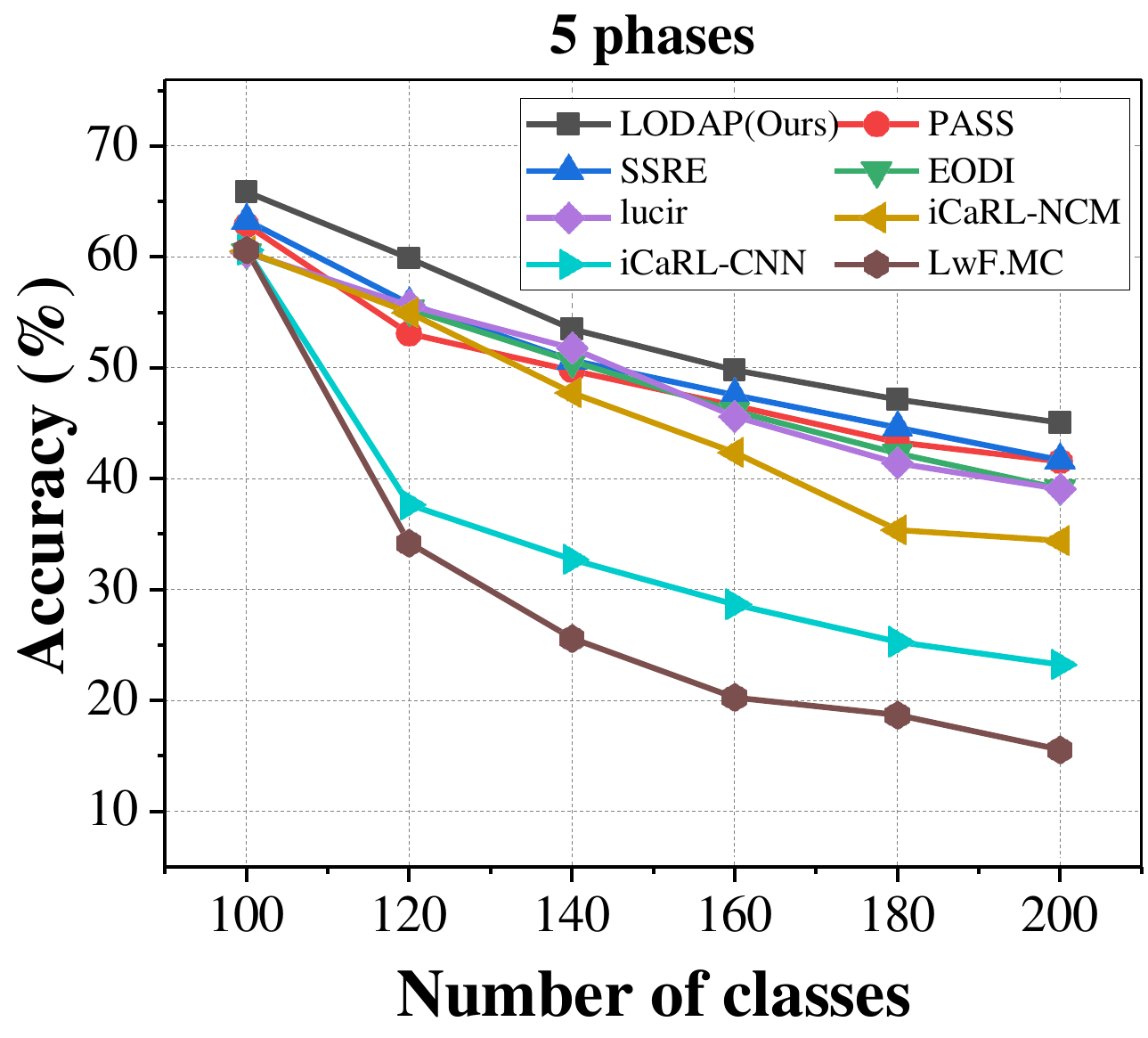}
        \label{5-tiny}
    \end{minipage}
    \begin{minipage}[t]{0.3\linewidth}
        \centering
        \includegraphics[width=\textwidth]{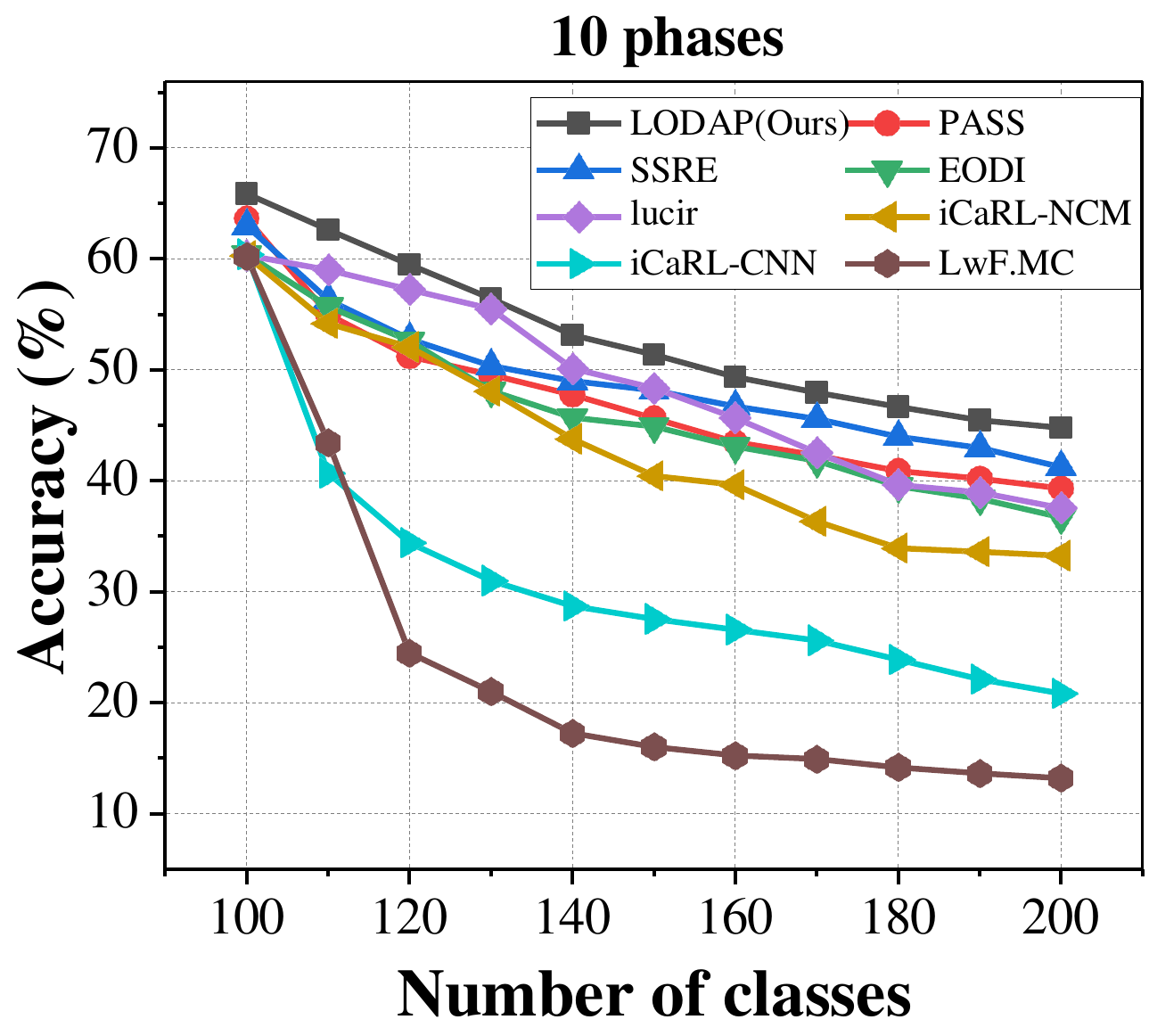}
        \label{10-tiny}
    \end{minipage}
        \begin{minipage}[t]{0.3\linewidth}
        \centering
        \includegraphics[width=\textwidth]{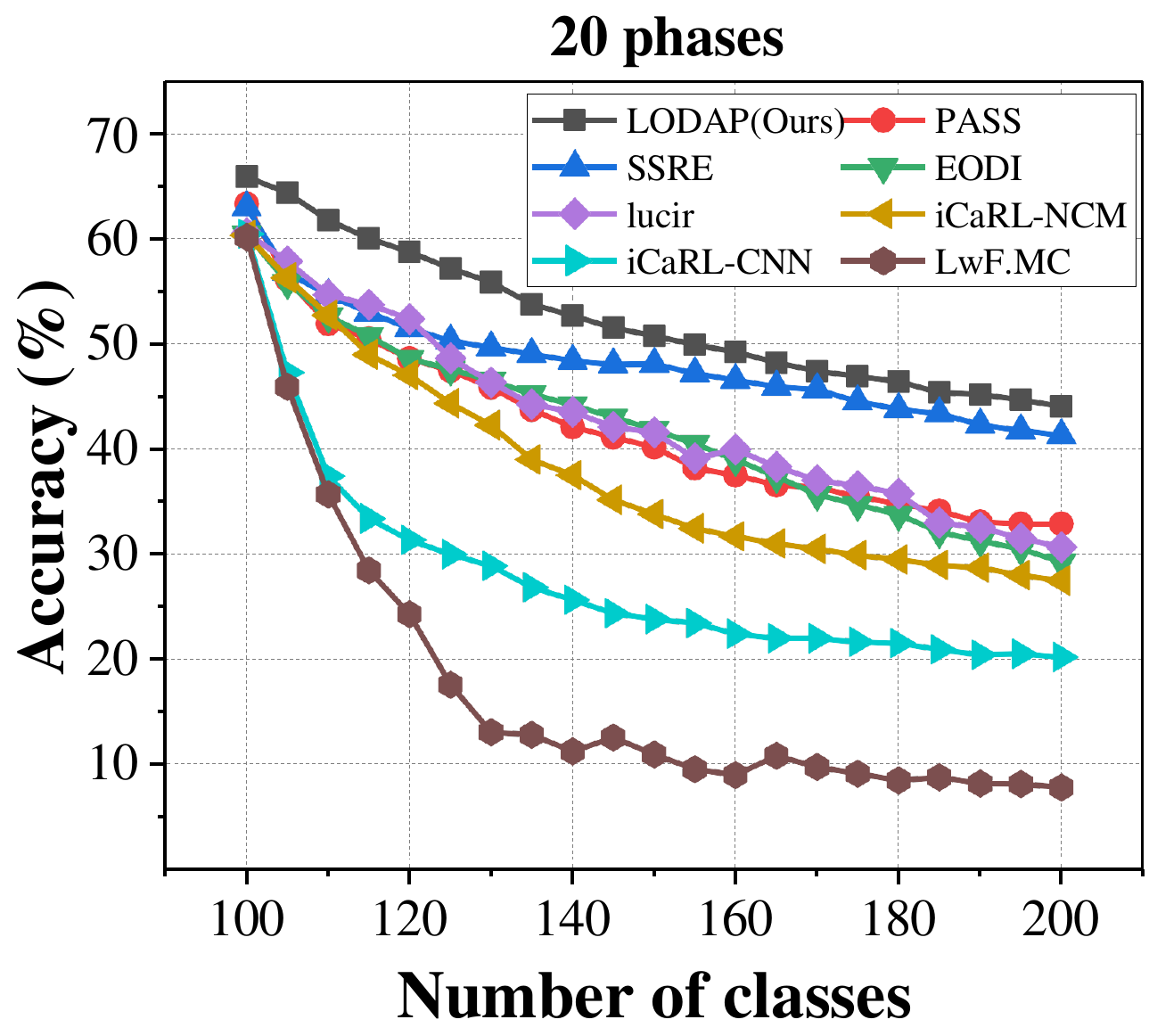}
        \label{20-tiny}
    \end{minipage}
	\caption{Classification accuracy at different incremental stages on the Tiny-ImageNet dataset.}
	\label{TinyImagenet}
\end{figure*}

\subsection{Evaluation}
\subsubsection*{Accuracy}
Table \ref{average accuracy} summarizes the accuracy results. 
\od obtains superior accuracy over the state-of-the-art (SOTA) methods, like \textsf{SSRE} \cite{zhu2022self}, \textsf{PASS} \cite{zhu2021prototype}, and \textsf{EODI} \cite{wang2022efficient}. 
Two results from \od are reported, one without data pruning (DP) and another with data pruning.
For CIFAR-100, \od increases the accuracy of 5 phases and 10 phases by 1.53\% and 1.29\%, respectively. Fig. \ref{CIFAR100} shows the accuracy variation during the course of IL.  
For Tiny-ImageNet, \od improves the accuracy by 3.17\%, 4.09\%, and 4.32\% at 5, 10, and 20 phase increments, respectively. Fig. \ref{TinyImagenet} shows the accuracy variation during the course of IL. 
Interestingly, \od with DP achieves consistent improvements across various datasets and incremental phases. Compared to the wo-DP approach, it achieves a maximum improvement of 0.36\%, while reducing the training samples by 70\%, significantly lowering the training cost.
From both Fig. \ref{CIFAR100} and \ref{TinyImagenet}, we can see that the forgetting phenomena still exists. 
As the number of classes increases, the model's accuracy gradually decreases. 
Nevertheless, \od always outperforms other methods in terms of retainable accuracy.

\subsubsection*{Model Complexity}
Thanks to the newly designed \eim, the respective numbers of parameters and FLOPs are significantly reduced, around 50\%, as shown in the first two columns of Table \ref{average accuracy}.
By means of adapter fusion, the complexity of the model remains the same during the IL procedure. The reduced model complexity may benefit the system in terms of power consumption. 
\subsubsection*{On-Device Performance}
\od features an on-device method, so in this section we show the training overhead and inference latency of different models on different platforms. 
Here, we demonstrate the overhead of training 10 new classes for one epoch. 
Table \ref{trainTime} shows the experimental results for training overhead. Here, we only show the overhead of five methods, including \od with and w/p DP, because other methods compared in Table \ref{average accuracy} have the same or higher training overhead than \textsf{SSRE}.
\textsf{PASS} has the highest training overhead because each image is augmented by three times for training. 
\od w/o DP has lower training overhead than \textsf{PASS} and \textsf{SSRE}. 
\textsf{EODI} has the lower training overhead than \od w/o DP because it deploys parameter freezing and learns fewer parameters during training. 
However, it suffers from the accuracy drop as seen in Table \ref{average accuracy}. 
\od w DP has the lowest training overhead because it prunes 70\% of training data.

Table \ref{inferenceTime} shows the inference latency of different methods. 
The latency is measured with batch size 1 and by the average value of 100 samples.
\textsf{PASS} and \textsf{EODI} show the same latency on all platforms because they do not change the model structure.  
Our \eim-modified model exhibits slightly higher latency than them on more powerful platforms, whereas our \eim-modified model is slightly faster than them on resource-constrained systems. 
The reason may be that the lightweight operations we use add more layers to the network. 
Since the model is executed layer by layer, the number of layers plays a more important role in model's inference. However, on less capable hardware, like Nano, both the model complexity and the number of layers have significant impact on a model's latency. \eim significantly reduce the model complexity, and this reduction remedies the increased latency caused by the more layers and finally translates to a latency improvement. 
Some techniques like \cite{fu2022depthshrinker} may reduce latency. 


\begin{table}[htbp]
    \centering
    \resizebox{0.97\linewidth}{!}{
    \begin{tabular}{|c|c|c|c|c|}
    \hline
        Method & RTX2060 & Jetson AGX Xavier & Jetson Nano \\ \hline
        \od w/o DP & 12.8s & 69.2s & 359.7s   \\ \hline
        \od w DP & 3.9s & 21.0s & 107.8s   \\ \hline
        \textsf{EODI} & 9.8s & 30.2s & 186.3s   \\ \hline
        \textsf{SSRE} & 15.9s & 74.1s & 390.1s   \\ \hline
        \textsf{PASS} & 53.4s & 193.1s &  1514.6s  \\ \hline
    \end{tabular}}
    \caption{Training overhead results on CIFAR-100. The batch size is 128, except for \textsf{PASS} due to its large memory occupation.}
    \label{trainTime}
\end{table}

\begin{table}[htbp]
    \centering
    \resizebox{0.97\linewidth}{!}{
    \begin{tabular}{|c|c|c|c|c|}
    \hline
        Method & RTX2060 & Jetson AGX Xavier & Jetson Nano \\ \hline
        \od & 4.64ms  & 8.5ms  & 21.0ms   \\ \hline
        \textsf{EODI} & 4.62ms  & 6.9ms  & 22.8ms   \\ \hline
        \textsf{SSRE} & 5.60ms  & 7.6ms  &  26.5ms  \\ \hline
        \textsf{PASS} & 4.54ms  & 6.4ms  & 23.2ms  \\ \hline
    \end{tabular}}
    \caption{The inference time of different methods with CIFAR-100 on different devices.}
    \label{inferenceTime}
\end{table}

\subsection{Ablation Study}
\subsubsection*{Loss Functions}
\begin{figure}[tb]
    \centering
    \includegraphics[width=0.65\columnwidth]{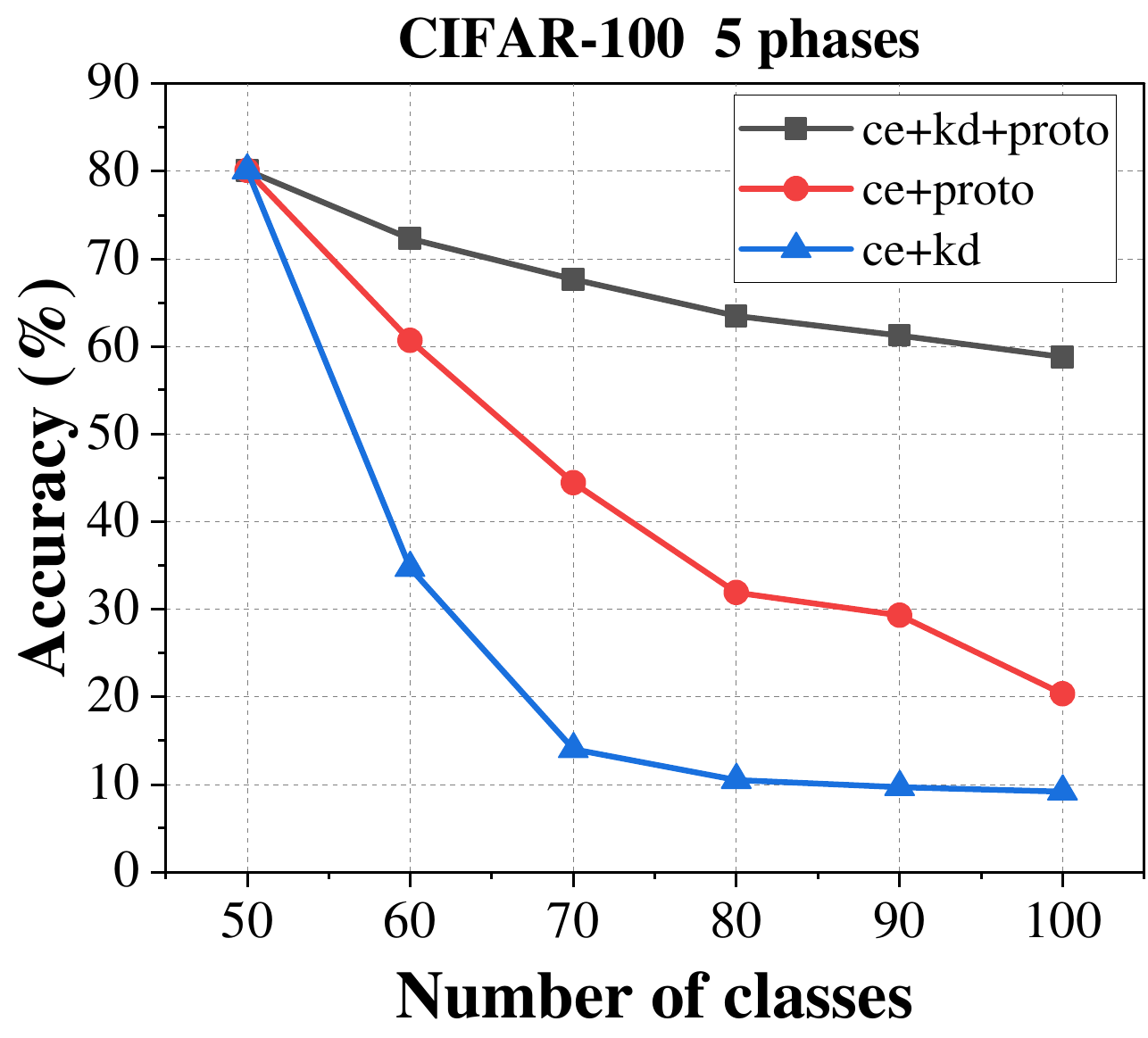}
	\caption{Ablation study of loss functions.}
	\label{loss}
\end{figure}

In order to assess the effectiveness of the three parts of our loss function, we conducted ablation experiments using the same settings of 5-phase experiment on CIFAR-100. 
The results are depicted in Fig. \ref{loss}. 
The experimental results show that only \textsf{ce+proto} degrades the average accuracy by 22.78\%. 
If only \textsf{ce+kd}, the average accuracy is significantly degraded by 40.85\%. 
The model trained with the three parts achieves the best accuracy. 
\subsubsection*{Data Pruning}
\begin{figure}[htbp]
    \centering
    \begin{subfigure}
        \centering
        \includegraphics[width=0.75\linewidth]{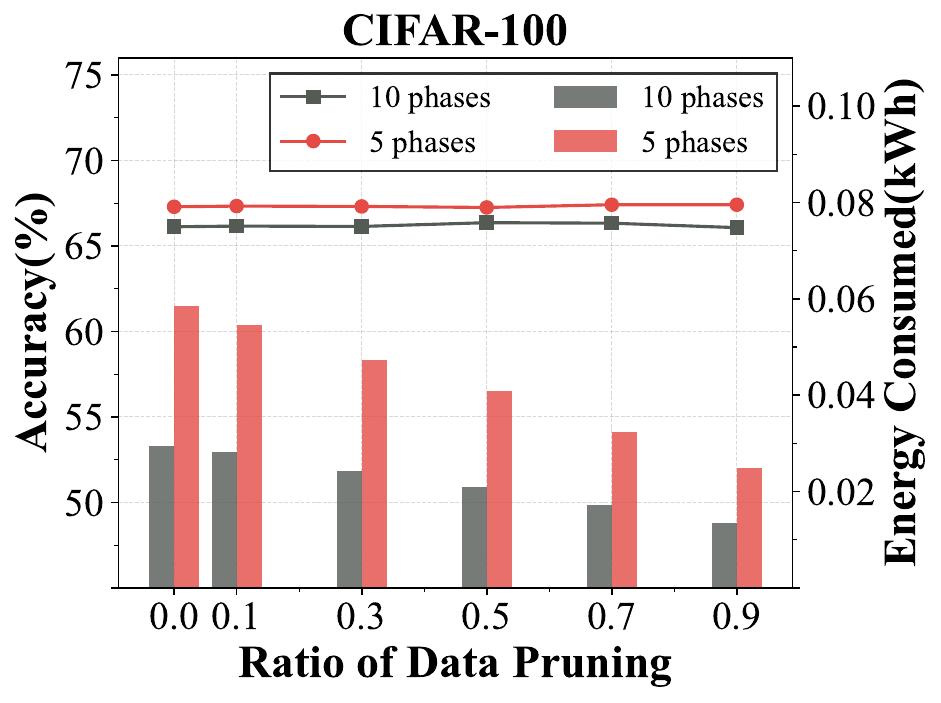}
        \label{cifar100}
    \end{subfigure}
    \begin{subfigure}
        \centering
        \includegraphics[width=0.8\linewidth]{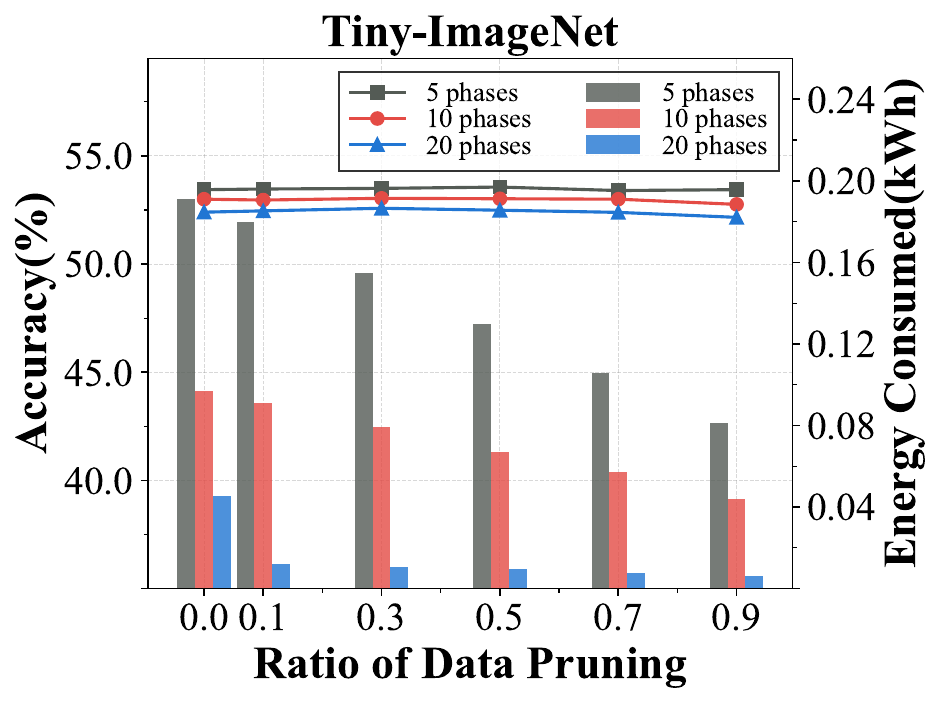}
        \label{Tiny-ImageNet}
    \end{subfigure}
    \caption{Ablation study of data pruning for CIFAR-100 and Tiny-ImageNet.}
    \label{DP}
\end{figure}

To evaluate the effectiveness of our data pruning strategy in maintaining model accuracy and reducing overhead, we conducted five incremental learning experiments across different datasets and incremental phases. For each experiment, we tested six pruning ratios while keeping all other training settings and optimization strategies consistent with those in Section \ref{exp_set}.  We used the same seed for each experiment with different pruning ratios. The experiments on the CIFAR-100 dataset were conducted on RTX2060, while those on the Tiny-ImageNet dataset were conducted on RTX4090. To evaluate both accuracy performance and training overhead, we recorded the following metrics:
\begin{itemize}
    \item \textbf{Time}: This is the total training time from the start of incremental learning to the end of the entire training process, that is, the total time for $p$ incremental phases.
    \item \textbf{Energy}: The average energy consumption during the training of each incremental phase.
    \item \textbf{Accuracy}: The average value of the model's test accuracies across all incremental steps.
    \item \textbf{Prototype saving time}: The time taken to save the prototypes at each incremental step.
\end{itemize}
The detailed results are shown in Table \ref{tab:DP-cifar100} and Table \ref{tab:DP-tinyimagenet}, while Figure \ref{DP} illustrates the accuracy and energy consumption for the experiments with different incremental steps on the two datasets. 
The lines show the accuracy results, while the bars visualize the energy consumption results. 
Taking the CIFAR-100 dataset as an example with 10 phases of incremental steps, after data pruning, the accuracy remains almost the same or even improves slightly compared to the ones without DP. 
The highest accuracy improvement is achieved by pruning 50\% of the data, where the average accuracy increases by 0.24\%. Additionally, the total training time was reduced by approximately 32\%, energy consumption decreased by about 28.8\%, and the time for prototype saving was also reduced by around 58.6\%.
For the CIFAR-100 dataset, under different pruning ratios from 0.1 to 0.9, the reduction in training time compared to training with the entire data ranges from 5.73\% to 57.47\%, while the reduction in energy consumption ranges from 4.07\% to 57.47\%. For the Tiny-ImageNet dataset, the largest accuracy improvement occurs in the 20-phase experiment, with a pruning ratio of 0.3, where the accuracy increases by 0.18\%. At this point, the training time is reduced by approximately 25.87\%, and energy consumption decreased by about 76.34\%. For the Tiny-ImageNet dataset with different pruning ratios, the reduction in training time ranged from 0.94\% to 85.72\%, and the reduction in energy consumption ranged from 5.74\% to 57.47\%.
Empirically, the optimal pruning ratio for CIFAR-100 is  0.7, while for Tiny-ImageNet, it is 0.5. 
This demonstrates that data pruning is an effective training optimization strategy, particularly suitable for scenarios with limited computational resources, such as edge systems.

\begin{table*}[htbp]
    \centering
    \small
    \begin{tabular}{c|c|c|c|c|c|c|c|c}
    \hline
         \multirow{2}{*}{DP ratio} & \multicolumn{4}{c|}{5 phases} & \multicolumn{4}{c}{10 phases} \\ \cline{2-9}  ~ & Time(s) & Energy(kWh) & Proto. time(s) & Acc.(\%) & Time (s)  & Energy (kWh) & Proto. time (s) & Acc. (\%)\\ \hline
         0 & 3549 & 5.84E-02 & 4.24 & 67.29 & 4829 & 2.95E-02 & 2.15 & 66.12 \\
         0.1 & 3346 & 5.45E-02 & 3.85 & 67.33 & 4545 & 2.83E-02 & 1.66 & 66.16 \\
         0.3 & 2945 & 4.73E-02 & 3.08 & 67.31 & 3950 & 2.43E-02 & 1.27 & 66.14 \\
         0.5 & 2507 & 4.08E-02 & 2.13 & 67.25 & 3283 & 2.10E-02 & 0.89 & 66.36 \\
         0.7 & 2009 & 3.23E-02 & 1.27 & 67.41 & 2672 & 1.72E-02 & 0.55 & 66.33 \\
         0.9 & 1549 & 2.48E-02 & 0.48 & 67.41 & 2050 & 1.35E-02 & 0.18 & 66.07 \\ \hline
    \end{tabular}
    \caption{Training cost and accuracy of the CIFAR-100 dataset at different incremental stages and data pruning ratios.}
    \label{tab:DP-cifar100}
\end{table*}

\begin{table*}[htbp]
    \centering
    \scriptsize
    \begin{tabular}{c|c|c|c|c|c|c|c|c|c|c|c|c}

    \hline
         \multirow{3}{*}{DP ratio} & \multicolumn{4}{c|}{5 phases} & \multicolumn{4}{c|}{10 phases} & \multicolumn{4}{c}{20 phases}\\ \cline{2-13} ~ & Time & Energy & Proto. time & Acc. & Time  & Energy & Proto. time & Acc. & Time & Energy & Proto. time & Acc. \\ ~ & (s) & (kWh) & (s) & (\%) & (s) & (kWh) & (s) & (\%) & (s) & (kWh) & (s) & (\%) \\ \hline
         0 & 8311 & 1.91E-01 & 13.69 & 53.44 & 8739 & 9.68E-02 & 6.58 & 53.00 & 8947 & 1.33E-02 & 3.13 & 52.40 \\
         0.1 & 7943 & 1.80E-01 & 12.89 & 53.47 & 8152 & 9.11E-02 & 5.71 & 52.96 & 8565 & 1.20E-02 & 2.95 & 52.40 \\
         0.3 & 6797 & 1.55E-01 & 9.43 & 53.47 & 6783 & 7.91E-02 & 4.42 & 52.96 & 7444 & 1.20E-02 & 2.20 & 52.58 \\
         0.5 & 5674 & 1.30E-01 & 6.85 & 53.56 & 5957 & 6.70E-02 & 3.18 & 53.02 & 6465 & 9.51E-03 & 1.77 & 52.49 \\
         0.7 & 4700 & 1.06E-01 & 4.33 & 53.40 & 5114 & 5.74E-02 & 1.88 & 53.00 & 5367 & 7.64E-03 & 1.02 & 52.39 \\
         0.9 & 3560 & 8.10E-02 & 1.35 & 53.44 & 3832 & 4.42E-02 & 0.65 & 52.76 & 4127 & 6.43E-03 & 0.36 & 52.16 \\ \hline
    \end{tabular}
    \caption{Training cost and accuracy of the Tiny-ImageNet dataset at different incremental stages and data pruning ratios.}
    \label{tab:DP-tinyimagenet}
\end{table*}

\label{exp:ablation}



\begin{figure}[tb]
    \centering
    \includegraphics[width=0.65\columnwidth]{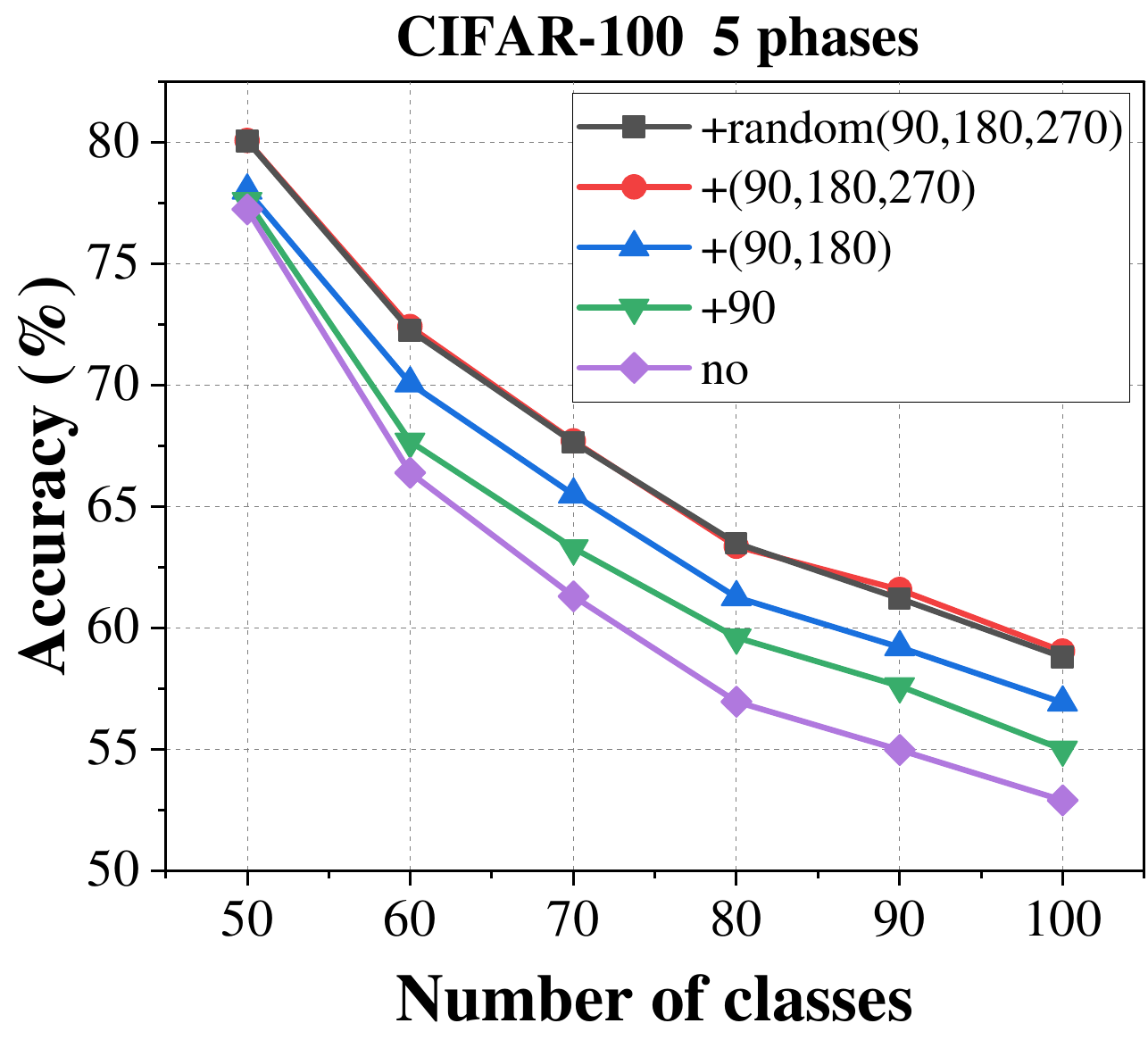}
	\caption{Ablation study of data augmentation strategies.}
	\label{feature_augmentation}
\end{figure}

\subsubsection*{Data Augmentation}
We also conducted an ablation study to access the effectiveness of our data augmentation strategy. 
As mentioned before, \textsf{PASS} applies data augmentation by rotating each input image by 90\degree, 180\degree, and 270\degree. 
Here, we compare 5 methods, \textit{no data augmentation}, \textit{only 90\degree}, \textit{90\degree and 180\degree}, \textit{all angels} (\textsf{PASS} exploits) and \textit{one randomly selected from the three angels} (the one used in \od). 
The results are shown in Fig. \ref{feature_augmentation}. 
Our data augmentation demonstrates almost the same accuracy as using all rotated images, but we only need half of the computation, 2 images vs. 4 images. 
Specifically, our method achieves accuracy improvements of 5.74\%, 3.91\%, and 2.20\% over no data augmentation, one rotation (90\degree), and two rotations (90\degree+180\degree), respectively. This demonstrates the effectiveness of data augmentation.


%% file: conclusion.tex
\section{Conclusion}
\label{sec:conc}
In this paper, we propose \od, a novel on-device incremental learning framework that combines the advantages of structure-based and regularization-based IL. \od introduces an efficient module for learning new features for newly added classes, while preserving the features of existing classes. It also employs data pruning strategies to reduce training costs on lightweight operations. Adapter fusion, the proposed data augmentation technique, and the corresponding training methods are integrated within \od. Extensive experiments provide strong evidence of \textsf{LODAP}’s effectiveness, efficiency, and applicability.


%% file: bare_conf.bbl
\begin{thebibliography}{10}
\providecommand{\url}[1]{#1}
\csname url@samestyle\endcsname
\providecommand{\newblock}{\relax}
\providecommand{\bibinfo}[2]{#2}
\providecommand{\BIBentrySTDinterwordspacing}{\spaceskip=0pt\relax}
\providecommand{\BIBentryALTinterwordstretchfactor}{4}
\providecommand{\BIBentryALTinterwordspacing}{\spaceskip=\fontdimen2\font plus
\BIBentryALTinterwordstretchfactor\fontdimen3\font minus
  \fontdimen4\font\relax}
\providecommand{\BIBforeignlanguage}[2]{{%
\expandafter\ifx\csname l@#1\endcsname\relax
\typeout{** WARNING: IEEEtran.bst: No hyphenation pattern has been}%
\typeout{** loaded for the language `#1'. Using the pattern for}%
\typeout{** the default language instead.}%
\else
\language=\csname l@#1\endcsname
\fi
#2}}
\providecommand{\BIBdecl}{\relax}
\BIBdecl

\bibitem{bengio2021deep}
Y.~Bengio, Y.~Lecun, and G.~Hinton, ``Deep learning for ai,''
  \emph{Communications of the ACM}, vol.~64, no.~7, pp. 58--65, 2021.

\bibitem{he2011incremental}
H.~He, S.~Chen, K.~Li, and X.~Xu, ``Incremental learning from stream data,''
  \emph{IEEE Transactions on Neural Networks}, vol.~22, no.~12, pp. 1901--1914,
  2011.

\bibitem{zhou2019edge}
Z.~Zhou, X.~Chen, E.~Li, L.~Zeng, K.~Luo, and J.~Zhang, ``Edge intelligence:
  Paving the last mile of artificial intelligence with edge computing,''
  \emph{Proceedings of the IEEE}, vol. 107, no.~8, pp. 1738--1762, 2019.

\bibitem{dhar2021survey}
S.~Dhar, J.~Guo, J.~Liu, S.~Tripathi, U.~Kurup, and M.~Shah, ``A survey of
  on-device machine learning: An algorithms and learning theory perspective,''
  \emph{ACM Transactions on Internet of Things}, vol.~2, no.~3, pp. 1--49,
  2021.

\bibitem{wang2022efficient}
Z.-H. Wang, Z.~He, H.~Fang, Y.-X. Huang, Y.~Sun, Y.~Yang, Z.-Y. Zhang, and
  D.~Liu, ``Efficient on-device incremental learning by weight freezing,'' in
  \emph{2022 27th Asia and South Pacific Design Automation Conference
  (ASP-DAC)}.\hskip 1em plus 0.5em minus 0.4em\relax IEEE, 2022, pp. 538--543.

\bibitem{de2021continual}
M.~De~Lange, R.~Aljundi, M.~Masana, S.~Parisot, X.~Jia, A.~Leonardis,
  G.~Slabaugh, and T.~Tuytelaars, ``A continual learning survey: Defying
  forgetting in classification tasks,'' \emph{IEEE transactions on pattern
  analysis and machine intelligence}, vol.~44, no.~7, pp. 3366--3385, 2021.

\bibitem{yan2021dynamically}
S.~Yan, J.~Xie, and X.~He, ``Der: Dynamically expandable representation for
  class incremental learning,'' in \emph{Proceedings of the IEEE/CVF conference
  on computer vision and pattern recognition}, 2021, pp. 3014--3023.

\bibitem{li2017learning}
Z.~Li and D.~Hoiem, ``Learning without forgetting,'' \emph{IEEE transactions on
  pattern analysis and machine intelligence}, vol.~40, no.~12, pp. 2935--2947,
  2017.

\bibitem{castro2018end}
F.~M. Castro, M.~J. Mar{\'\i}n-Jim{\'e}nez, N.~Guil, C.~Schmid, and K.~Alahari,
  ``End-to-end incremental learning,'' in \emph{Proceedings of the European
  conference on computer vision (ECCV)}, 2018, pp. 233--248.

\bibitem{paul2021deep}
M.~Paul, S.~Ganguli, and G.~K. Dziugaite, ``Deep learning on a data diet:
  Finding important examples early in training,'' \emph{Advances in neural
  information processing systems}, vol.~34, pp. 20\,596--20\,607, 2021.

\bibitem{rebuffi2017icarl}
S.-A. Rebuffi, A.~Kolesnikov, G.~Sperl, and C.~H. Lampert, ``icarl: Incremental
  classifier and representation learning,'' in \emph{Proceedings of the IEEE
  conference on Computer Vision and Pattern Recognition}, 2017, pp. 2001--2010.

\bibitem{bang2021rainbow}
J.~Bang, H.~Kim, Y.~Yoo, J.-W. Ha, and J.~Choi, ``Rainbow memory: Continual
  learning with a memory of diverse samples,'' in \emph{Proceedings of the
  IEEE/CVF conference on computer vision and pattern recognition}, 2021, pp.
  8218--8227.

\bibitem{douillard2020podnet}
A.~Douillard, M.~Cord, C.~Ollion, T.~Robert, and E.~Valle, ``Podnet: Pooled
  outputs distillation for small-tasks incremental learning,'' in
  \emph{Computer vision--ECCV 2020: 16th European conference, Glasgow, UK,
  August 23--28, 2020, proceedings, part XX 16}.\hskip 1em plus 0.5em minus
  0.4em\relax Springer, 2020, pp. 86--102.

\bibitem{hou2019learning}
S.~Hou, X.~Pan, C.~C. Loy, Z.~Wang, and D.~Lin, ``Learning a unified classifier
  incrementally via rebalancing,'' in \emph{Proceedings of the IEEE/CVF
  conference on computer vision and pattern recognition}, 2019, pp. 831--839.

\bibitem{fernando2017pathnet}
C.~Fernando, D.~Banarse, C.~Blundell, Y.~Zwols, D.~Ha, A.~A. Rusu, A.~Pritzel,
  and D.~Wierstra, ``Pathnet: Evolution channels gradient descent in super
  neural networks,'' \emph{arXiv preprint arXiv:1701.08734}, 2017.

\bibitem{kirkpatrick2017overcoming}
J.~Kirkpatrick, R.~Pascanu, N.~Rabinowitz, J.~Veness, G.~Desjardins, A.~A.
  Rusu, K.~Milan, J.~Quan, T.~Ramalho, A.~Grabska-Barwinska \emph{et~al.},
  ``Overcoming catastrophic forgetting in neural networks,'' \emph{Proceedings
  of the national academy of sciences}, vol. 114, no.~13, pp. 3521--3526, 2017.

\bibitem{zhu2022self}
K.~Zhu, W.~Zhai, Y.~Cao, J.~Luo, and Z.-J. Zha, ``Self-sustaining
  representation expansion for non-exemplar class-incremental learning,'' in
  \emph{Proceedings of the IEEE/CVF Conference on Computer Vision and Pattern
  Recognition}, 2022, pp. 9296--9305.

\bibitem{zhu2021prototype}
F.~Zhu, X.-Y. Zhang, C.~Wang, F.~Yin, and C.-L. Liu, ``Prototype augmentation
  and self-supervision for incremental learning,'' in \emph{Proceedings of the
  IEEE/CVF Conference on Computer Vision and Pattern Recognition}, 2021, pp.
  5871--5880.

\bibitem{toneva2018empirical}
M.~Toneva, A.~Sordoni, R.~T.~d. Combes, A.~Trischler, Y.~Bengio, and G.~J.
  Gordon, ``An empirical study of example forgetting during deep neural network
  learning,'' \emph{arXiv preprint arXiv:1812.05159}, 2018.

\bibitem{coleman2019selection}
C.~Coleman, C.~Yeh, S.~Mussmann, B.~Mirzasoleiman, P.~Bailis, P.~Liang,
  J.~Leskovec, and M.~Zaharia, ``Selection via proxy: Efficient data selection
  for deep learning,'' \emph{arXiv preprint arXiv:1906.11829}, 2019.

\bibitem{mirzasoleiman2020coresets}
B.~Mirzasoleiman, J.~Bilmes, and J.~Leskovec, ``Coresets for data-efficient
  training of machine learning models,'' in \emph{International Conference on
  Machine Learning}.\hskip 1em plus 0.5em minus 0.4em\relax PMLR, 2020, pp.
  6950--6960.

\bibitem{he2024you}
Y.~He, L.~Xiao, and J.~T. Zhou, ``You only condense once: Two rules for pruning
  condensed datasets,'' \emph{Advances in Neural Information Processing
  Systems}, vol.~36, 2024.

\bibitem{han2020ghostnet}
K.~Han, Y.~Wang, Q.~Tian, J.~Guo, C.~Xu, and C.~Xu, ``Ghostnet: More features
  from cheap operations,'' in \emph{Proceedings of the IEEE/CVF conference on
  computer vision and pattern recognition}, 2020, pp. 1580--1589.

\bibitem{ding2021repvgg}
X.~Ding, X.~Zhang, N.~Ma, J.~Han, G.~Ding, and J.~Sun, ``Repvgg: Making
  vgg-style convnets great again,'' in \emph{Proceedings of the IEEE/CVF
  conference on computer vision and pattern recognition}, 2021, pp.
  13\,733--13\,742.

\bibitem{fu2022depthshrinker}
Y.~Fu, H.~Yang, J.~Yuan, M.~Li, C.~Wan, R.~Krishnamoorthi, V.~Chandra, and
  Y.~Lin, ``Depthshrinker: a new compression paradigm towards boosting
  real-hardware efficiency of compact neural networks,'' in \emph{International
  Conference on Machine Learning}.\hskip 1em plus 0.5em minus 0.4em\relax PMLR,
  2022, pp. 6849--6862.

\end{thebibliography}
